# An end-to-end, interactive Deep Learning based Annotation system for cursive and print English handwritten text

Pranav Guruprasad[1], Sujith Kumar S[2], Vigneswaran C[2], and V. Srinivasa Chakravarthy[2]

[1]Birla Institute of Technology and Science Pilani, K. K. Birla Goa Campus, India
[2]Bhupat and Jyoti Mehta School of Biosciences, Department of Biotechnology, Indian Institute of Technology, Madras, Chennai, India

f20171918@goa.bits-pilani.ac.in[1], schakra@ee.iitm.ac.in[2]

**Abstract** With the surging inclination towards carrying out tasks on computational devices and digital mediums, any method that converts a task that was previously carried out manually, to a digitized version, is always welcome. Irrespective of the various documentation tasks that can be done online today, there are still many applications and domains where handwritten text is inevitable, which makes the digitization of handwritten documents a very essential task. Over the past decades, there has been extensive research on offline handwritten text recognition. In the recent past, most of these attempts have shifted to Machine learning and Deep learning based approaches. In order to design more complex and deeper networks, and ensure stellar performances, it is essential to have larger quantities of annotated data. Most of the databases present for offline handwritten text recognition today, have either been manually annotated or semi automatically annotated with a lot of manual involvement. These processes are very time consuming and prone to human errors. To tackle this problem, we present an innovative, complete end-to-end pipeline, that annotates offline handwritten manuscripts written in both print and cursive English, using Deep Learning and User Interaction techniques. This novel method, which involves an architectural combination of a detection system built upon a state-of-the-art text detection model, and a custom made Deep Learning model for the recognition system, is combined with an easy-to-use interactive interface, aiming to improve the accuracy of the detection, segmentation, serialization and recognition phases, in order to ensure high quality annotated data with minimal human interaction.

# 1 Introduction

Handwriting Recognition has been a field of extensive research, for the past few decades, and has evolved from approaches which employ the likes of Hidden Markov models [1], to Deep learning approaches [2-3] in the recent past. Optical Character Recognition (OCR) systems that carry out the task of complete digitization, including text detection and recognition, have also been prevalent for a long time. Even though OCR systems have come a long way and produce excellent results on print text, their success has not carried over as much to cursive handwritten text digitization and face various challenges [4]. This can be attributed mainly to the variations in styles of handwriting, the spacing, lighting, geometrical orientations, noise, problems in segmentation, and much more. Even though many OCR systems of late are using Deep Learning approaches since the boom in the Deep Learning era, the results of these have been impressive only on handwritten documents with certain predefined structural formats, and not on simple, common handwritten documents and manuscripts. These drawbacks are because most OCR implementations are based on templating and feature extraction techniques. Thus, it is clear that there is still abundant room for improvements in the field of handwritten text recognition and digitization. To aid further research in these fields, we propose an annotation system for cursive and print handwritten English text, that provides fast and state-of-the-art quality annotated data while requiring hardly any human intervention or effort. Our annotation system offers a big advantage for users to create a comprehensive annotated dataset on any kind of handwritten documents of their choice, with no prerequisites for specified styles, formats, appearances or cumbersome preprocessing techniques. Upon passing mere photographs or scanned copies of the individual pages of a handwritten document into our pipeline, it converts them into a coherent annotated dataset that the user can use further for any desired tasks. Our pipeline consists of a word detection system built upon the state-of-the-art text detection model - EAST; an interactive user interface built using Python TKinter, that provides the user an opportunity to remove the common errors in the detection, segmentation and serialization phases which are committed by even the best text detectors; and finally a powerful Deep Learning model for the recognition phase, which boasts of a custom designed multi-dimensional LSTM (Long Short Term Memory units), Convolutional Neural Network (CNN) and a Connectionist Temporal Classifier(CTC). Even though we provide word level annotations, the recognition is implemented on a character level, thus allowing the recognition system to recognize words beyond its training data. In addition to this, meticulous and innovative data preprocessing techniques have been implemented on the images of the words that are detected and passed on to the recognition system, which helps bolster the robustness of the system and increase the accuracy of the annotations.

## 2 Related Works

The advent of Deep Learning and its rise in popularity, led to an increased need for annotated data for supervised learning tasks. Most of the early datasets have been annotated manually, and only in the recent past has there been attempts to automate the process of annotation and reduce the human efforts involved in it. However, it is important to keep in mind that in annotation, feedback from the user is indispensable in some cases in order to create the highest quality of annotated data that is nearly devoid of any errors.

There have been attempts to automate the annotation of data in various fields such as real time video feeds [5], object detection [6], and even the semantic web [7] . Similarly, with the evident need for annotated data for offline handwritten text, especially with the shift towards Deep Learning based approaches for offline handwriting text recognition, it is not surprising that there has been a lot of research in this area too. There have been quite a few works that have proposed a systematic arrangement of stages to create a complete annotation engine for handwritten text, comprising of varying levels of automation [8-9]. However, a complete end-to-end pipeline to annotate handwritten text with very minimal human interaction is still considered to be a very challenging task as discussed in a part of a study by Ung et al. [10].

The two main components of our annotation pipeline are a word detection system and a handwriting recognition system. There has been extensive research in these fields for the past many years. There have been non-Machine Learning approaches, Machine Learning approaches, and most recently, Deep Learning approaches. Lavrenko et al. [11] presented a Hidden Markov Model based holistic word recognition approach, inspired by the results in cognitive psychology, for word recognition in handwritten historical documents. This method, which did not implement the segmentation of words into characters, gave a recognition accuracy of 65% which exceeded the results of the other systems during that time. Optical Character Recognition (OCR) systems have been around for a very long time, and there have been many attempts for Handwritten OCR, for various languages [12]. OCR techniques have evolved quantifiably over this period, and over the past few years with the advent of cloud computing, GPUs, and a better research community, have shifted towards some very impressive Deep Learning based models [13-15]. However, as mentioned in the previous section, OCR techniques face various challenges [16] and have not been able to provide exciting results for cursive handwritten text documents that lack a predefined structure.

A work done by Shiedl et al. [17] implements a handwriting recognition system, with an architecture based on Convolutional Neural Networks (CNN), Recurrent Neural Networks (RNN) and a Connectionist Temporal Classifier (CTC), which provides impressive results for handwriting recognition. This was one of the main inspirations for the recognition architecture we present in this work. The 2D LSTM implemented in the recognition system of our work, is inspired by a work by Graves et al. [18].

Text detection has also been an extensively researched field, especially with the advancements in image processing, object detection and deep learning. Many works follow different categories of techniques for scene text detection and document text detection, like the sliding window techniques[19], single shot detection techniques[20] and region based text detection techniques [21]. Zhou et al. [22] proposed a simple yet robust deep learning based scene text detector known as EAST - Efficient and Accurate Scene Text Detector. In our work we build upon this EAST model as it not only outperforms a majority of the state-of-the-art text detectors in terms of accuracy and speed, but is easy to build upon, and works satisfactorily on different orientations of text words which provides a great advantage when dealing with handwritten text detection, given the variability in handwriting style from person to person.

## 3 Data and Preprocessing

This section mentions the datasets used for training and testing the detection and recognition architectures in our pipeline, and also discusses the various preprocessing techniques applied on the words, both before training the recognition system, and also before testing or implementing the recognition system on the unseen data, in order to help increase accuracy of recognition.

### 3.1. Data

The EAST model upon which our detection system is built, was trained on the ICDAR 2013 [23] and 2015 datasets [24]. The ICDAR 2015 dataset consists of a total of 1500 scene text images, out of which 1000 are training data and the remaining are test data. The ICDAR 2013 dataset contains 229 training images which were also used additionally for training. For the recognition system, our model was trained on the entire IAM Offline Handwriting Database [25]. This dataset contains words in both print and cursive handwritten English text that have been written by 657 writers. These pages have been scanned, automatically segmented, and then manually verified. Containing over 1500 pages of scanned text and over 1,00,000 labeled words, this dataset provides the massive volume of data that would be required to train the deep network of our recognition system, in order to perform well. Apart from this, the dataset offers a large diversity in the various features of offline handwritten text such as style, geometric orientation and spacing, thus ensuring the robustness of the handwriting recognition system. This enhances the performance of the system not only with respect to new, unseen data, but also transfer learning. Apart from the IAM offline dataset, we used the CVL dataset [26] to test the robustness of our recognition system, as the handwriting styles in this dataset make it very hard to recognize the words and are not considered to be very legible .

## 3.2. Preprocessing

### 3.2.1 For training the recognition system

The IAM dataset consists of grayscale images of individual words. There is a variation in width and height of the images due to the lengths of different words and the heights of different characters. To maintain uniformity, and to make it easier to pass the images as inputs to the model, each image was resized such that it has the width or height of at least 1, and either a width of 128, or height of 32 at most . This resized input image was then copied into a complete white image which had a fixed width of 128 and height of 32 (128x32). Thus, every input had a uniform dimension of 128x32 without getting distorted in the process. Basic data augmentation techniques were used to increase the dataset size and make it robust to common variations that occur in the dataset. As a part of this, the input images were subject to random stretches, for which the stretch values were obtained by a random function set within a specified range. They were also shifted horizontally and vertically by small amounts so that even if parts of a character were cropped out, the recognition system would still be able to recognize the whole word. After these augmentations, the grayscale values of the images were also normalized, just to make the task easier for the network. Some of the images in the IAM dataset are damaged, and to take care of this problem, black images of the same dimensions were used in place of these damaged files. In addition to all these techniques, during the training process, the recognition system was additionally trained on the same images but with an added randomized Gaussian noise which helped reduce any chances of overfitting the IAM dataset, and enabled the recognition system to recognize the word to a good accuracy level irrespective of the gray values and contamination of the new image due to various natural causes.

### 3.2.2 Processing applied to the new unseen data

**Converting to the IAM format**

The pictures or scanned copies of the handwritten text documents that have to be annotated, may look very different from the images used to train the recognition system. The neural network not only learns how to read the words but also learns features like the contrast, the thickness of the words, the style and even learns the features of the surroundings.

To make sure that these factors do not affect the results, and to cater to a huge variety of handwritten text in documents, we process the word images (which are passed on in the pipeline to the recognition system) that do not resemble the images in the IAM dataset. To make sure they are very similar to the IAM images, three functions are carried out :

1. The contrast of the images is increased to a high contrast level
2. The word images are cropped to a very tight fit around the text
3. The thickness of the text is increased to make it resemble a bold font style

Figure 1(a) shows a handwritten image that is initially not in the IAM data format, and Figure 1(b) shows how the preprocessing step converts it to resemble IAM dataset style images.

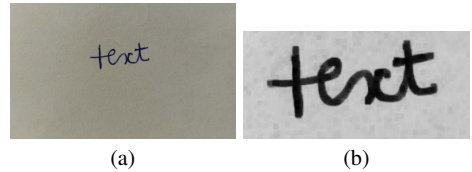

(a)  (b)

**Fig. 1** Handwritten text image before and after the preprocessing steps to convert to IAM style text (a) Non IAM handwritten text image before preprocessing (b) Same handwritten text image as (a), but after the preprocessing step done to convert to IAM style image

This preprocessing step has been observed to increase the accuracy of recognition significantly for word images that look different from the training data.

**Removing the slope and slant (Normalizing)**

The amount of slope and slant in cursive English words varies from writer to writer. Since this change in slope and slant affects the appearance and geometric orientation of the characters, this variation may have a significant effect on the performance of the recognition system.

To make sure that irrespective of the writer's style, the recognition system accurately recognizes the characters, while testing on the unseen data - we use a normalization technique to remove the slope and slant to a fair extent from all the words that come through the pipeline into the recognition system, so that they all have roughly the same amount of slope and slant. This makes sure that a cursive character written by two writers with completely different styles, would still look very similar to each other and thus prevent the recognition system from making any errors. When an input image is passed into the function that carries out this operation to remove the slant,the entire text part of the image is deslanted and made fairly upright, and the empty part of the image is filled with white colour. The algorithm we use for this slope and slant removal is based on a work by Vinciarelli et al. [27], and was implemented using OpenCV. Figure 2(a) shows an image of handwritten text with slant and slope, and Figure 2(b) shows the same handwritten text after the slope and slant have been removed.

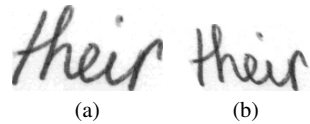
(a)   (b)

**Fig. 2** Handwritten text image before and after the slope and slant removal (a) A handwritten text image from the IAM dataset (b) Same handwritten text in (a), but after slant and slope removal

## 4 Methodology and functioning

**Phases**

The functionality of the proposed pipeline can be divided into independent phases, with each phase contributing sequentially, towards the goal of completely annotating the document. In this section we discuss the functioning of the pipeline, phase by phase when the photo or scanned copy of a single page of the handwritten text document is passed as input. Figure 6 shows a simple overview of the sequence of phases in our pipeline.

### 4.1. Word detection

The first phase in the pipeline is the detection of each of the words in the photograph of the page uploaded.Our detection model was built upon the pre-trained EAST model, which is a very robust text detector and achieves state-of-the-art text detection accuracies such as an F-score of 0.7820 on the ICDAR 2015 dataset. The page that is uploaded to our pipeline, is resized to a standard 720p image, before passing it into the detection system for detecting the individual words. For a page of handwritten text , our detection system which boasts of very impressive speed, produces its final set of bounding boxes around all the words it detects in the page, in just around an average of 0.5 seconds. In this network, one output layer gives the coordinates of all the initial bounding boxes that are predicted for the words detected in the page. In addition to this output layer, there is another output layer with a sigmoid activation function that outputs the probabilities signifying the presence of text in different regions of the image. After these values are obtained, a Non-Maximum Suppression technique is implemented to remove the weak and overlapping bounding boxes which are associated with lesser probability scores (which do not have a higher probability than a specified threshold). This then results in the final set of the bounding boxes predicted by the system, with their respective coordinates, for all the words it could detect in the page. After this function is complete, there are still some words that are undetected (which are usually one or two lettered words), some wrongly located bounding boxes, and some bounding boxes that do not cover the entire word vertically, or horizontally, or along both directions. These are common mistakes committed even by state-of-the-art object and text detection systems. Especially in our case, since it is handwritten text, the data is rife with noise and style variations, which only increases the chances of these detection imperfections. To make sure that we do not miss out on annotating every single word and symbol in the document, and to ensure that the accuracy levels of the recognized words which are used to annotate the document are kept high, the pipeline leads into the next phase, which is the interactive interface that allows the user to intervene and rectify these imperfections with minimal effort.

### 4.2. Interactive Interface

#### 4.2.1 Editing, Adding and Deleting Bounding Boxes

As mentioned in the previous section, there are always certain cases where human intervention is inevitably required to create near flawless annotations, which is facilitated in our work by this interactive interface. The interface was completely built using Tkinter, which is a Python interface to the Tk GUI toolkit. Tkinter is widely used as a standard GUI for Python implementations, and works across all the popular operating systems.

This part of the pipeline first displays the entire image of the input handwritten page after being resized, on a Tkinter canvas of fixed size, for the user to see. Then the bounding box coordinates from the detection system of the pipeline are retrieved, and scaled according to the Tkinter canvas size. Once the coordinates are obtained with respect to the canvas size, these bounding boxes are displayed as interactive red rectangles at the coordinates where the detection system predicted the bounding boxes on the page. Before the serialization of words, this interface provides 4 main functionalities - Adding new bounding boxes, deleting bounding boxes, resizing bounding boxes and moving around of bounding boxes.

Adding Bounding Boxes: This feature can be used by the user when the detection system has not recognized certain small words or punctuation symbols. Usually text detection systems do not recognize one or two lettered words that are written so shabbily that they hardly look like text, or some punctuation symbols like full stops or commas, that are written too lightly and are barely recognizable as a written component. This can be corrected by drawing a bounding box over any symbol or word that has not been detected. The drawing of the bounding box is made very easy for the user and a left click and drag of the mouse anywhere on the canvas simply draws a bounding box, and this is logged into the system automatically as a new bounding box, along with its respective details such as coordinates.

Deleting Bounding Boxes: This feature comes in handy when the system has detected wrong spaces of the image as a text containing region, or if there are two bounding boxes over one hyphenated words, and many other scenarios like this. We allow the user to remove any bounding box on the canvas with the simple press of a key, after which the information stored regarding this bounding box is deleted in the system.

Resizing/moving Bounding Boxes: These features are useful when the bounding box coordinates detected by the detection system are not entirely accurate and do not cover the entire part of the word or symbol. This is a very common problem in the already existing methods and can affect the output of the recognition system adversely. The resizing functionality that we provide, allows the user to resize bounding boxes very easily, identical to how one would resize an image in a word document or a powerpoint presentation document, by dragging along the corners or edges of the bounding box. Moving the bounding boxes around as a whole, can be done by hovering the mouse over the desired bounding box and pressing any one of the arrow keys according to the required direction. Once the bounding boxes have been shifted or resized, their respective modified coordinates are automatically stored. Once these editing operations are carried out and the user is satisfied with the positions of the bounding boxes around all the words and symbols in the page, the serialization of the words can be carried out. Figure 3 shows examples of when these operations are required and how our interface allows the user to fix them.

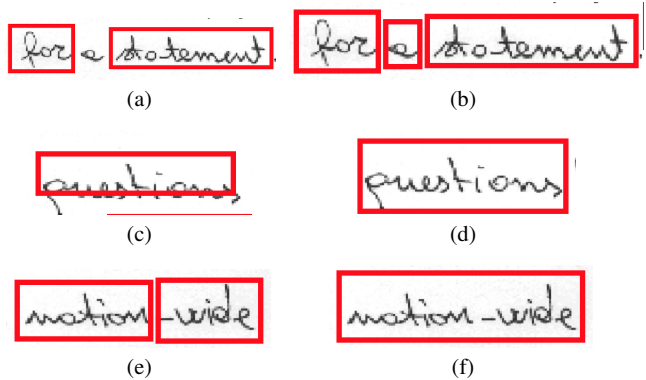

**Fig. 3** (a) The detection system missed the detection of a one lettered word - "a" (b) The interface allows the user to draw a bounding box so that "a" is detected (c) The detection system does not draw the bounding box accurately enough to enclose the whole word (d) The interface allows the user to resize the bounding box so that the whole word is enclosed (e) The detection system detects one hyphenated word as two separate Words (f) The interface allows the user to delete the extra box, move around and resize the other box so that it is enclosed as one

### 4.2.2 Serialization

The serialization phase is a very important phase as it determines the order of the detected words in the page of the document. Once the editing phase is complete and verified by the user, just a simple press of a specific key, automatically serializes all the words and symbols that have bounding boxes around them, and is represented on the canvas by straight lines between the bounding boxes to indicate their order, as it can be seen in Figure 4. The serialization function was implemented by our own sorting algorithm that was based on the scaled canvas coordinates of the bounding boxes, the resolution of the image, the space between the lines of text and the space between the adjacent words in a line. Once this serialization is complete and is visible to the user, the user can further modify it in case they desire to change the order of any of the words. We have enabled a swapping feature to carry this task out, which we facilitated by constructing an elaborate dictionary data structure that contains information regarding the coordinates of each of the bounding boxes, their neighbours on either sides based on the serialized order, and their object tags in the canvas. By just right clicking on two bounding boxes and pressing a key, the serialized order of the two bounding boxes are swapped. This change is not only shown on the interface, but also automatically logged in the system, and the new final serialized order is stored.

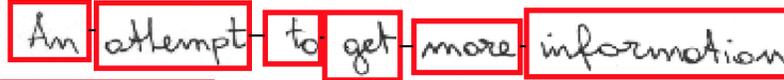

**Fig. 4** The black solid lines between the bounding boxes indicate the order in which the words have been serialized

### 4.3. Recognition

Once the serialization phase is complete, the order of the words present in the document is now finalized. The images of these individual words, within each of the bounding boxes, are then extracted and stored using OpenCV. They are then passed on, one by one (in the same serialized order), as individual word inputs to the recognition system in the pipeline. Each input image that is passed into the recognition system goes through the three architectures of our recognition model : the Convolutional Neural Network (CNN), the multi-dimensional LSTM, and the Connectionist Temporal Classifier(CTC). The exact details and nature of these architectures are discussed in the next section.

For each input of an individual image to the recognition system, the output is a sequence of characters. The recognition system thus outputs a sequence of characters/ individual character, for each of the detected, and serialized images passed into it from the previous phases of the pipeline. As the image passes through each of the layers of the CNN, the trained layers extract all the required features from that image. There are three main operations that are carried out on the image in the CNN, in each layer: the convolutional operation, a non-linear activation and a pooling function. Apart from these three operations, we add a Gaussian noise layer that adds standard Gaussian noise to the input, for reasons explained in 3.2.1. Then finally, after passing through two fully connected layers, a feature map is output.

The output feature map from the CNN is then passed as input to our 2D LSTM. An LSTM is used instead of a standard unidirectional or bidirectional RNN, as LSTMs prevent loss of information over long distances, and so is very helpful when dealing with long character sequences, which is very important for the task at hand. Our custom 2D LSTM was designed and implemented instead of using a standard one dimensional LSTM, because we felt that considering both the horizontal and vertical dimensions of handwritten text while recognizing it, would be much more effective than just working along one dimension. This is because the English cursive handwritten text has a myriad of variations along both the dimensions, and the system would be very robust if it could learn the features across both these dimensions. Moysset et al. [28] show that 2D LSTMs give great results for recognition of handwritten text, and provide higher performances as compared to single dimensional RNNs or LSTMs even when used on complex, challenging and real life data.

The output sequence of the LSTM is mapped to a matrix which becomes the input to the final CTC layer. Connectionist Temporal classification, a work by Graves et al. [29] is a method that serves two purposes : it not only calculates the loss values required for training, but also decodes the matrix that is output from the LSTM, to obtain the final text that is present in the input image. During the training process, both the ground truth and the LSTM output matrix are fed to the CTC layer, and based on these, a loss value is calculated which is used to train the system to recognize the right sequence of characters. During the inference phase, only the LSTM output matrix is fed, and is decoded by the CTC layer, using a decoding algorithm, to get the text from the images. As the text from each input image is recognized by the recognition system, they are checked for any misspells, and are corrected using a state-of-the-art python spell checker known as Pyspellchecker. This spell checker which was developed and released very recently, even offers a feature where the users can add words of their choice to the dictionary, thus allowing them customize the dictionary to suit the task at hand. After this stage, all the recognized words and symbols are stored in the same serialized order. Then, all these stored texts are retrieved and displayed in text boxes on the Tkinter canvas, under each of their corresponding detected words/symbols, as the annotations for that word/symbol, as seen in Figure 5. Then we offer a feature that allows the user to interact again with the pipeline before the final annotations are stored. The user is allowed to edit the recognized text present in each of the text boxes according to their desire. This ensures that any small mistakes made by the system are corrected completely before the annotations are finalized. After this stage, the user can click a button that shows up on the canvas, upon which the final modified annotations for all the words are stored and written into a text file in the same order as they appear in the original handwritten document, and are accessible to the user.

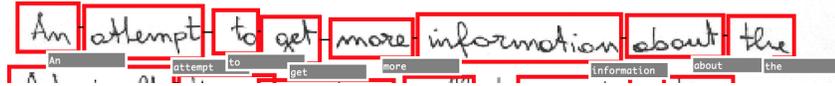

**Fig. 5** The recognized words are displayed in editable text boxes below their corresponding detected words

### 4.4. Details of the Recognition model

The combination of a CNN, RNN and a CTC layer has been gaining popularity in the recent past, especially for text recognition tasks. We modify this architectural combination, and improve upon it by building our own 2 Dimensional LSTM to replace the standard RNN, which results in a significant increase in performance. Apart from this, the CNN in our system has also been designed in a way such that it is powerful and robust enough to deal with handwritten text images. The CNN model contains 10 layers, out of which each of the first 8 layers perform convolutional operations, non linear activations and pooling functions. The convolutional operations are carried out by filters of varying kernel sizes from 7x7 to 3x3, and a standard RELU non linear activation function is used. These 8 layers are followed by a Gaussian noise layer and 2 fully connected layers. The input to the CNN is the preprocessed image of dimensions 128x32, and the output is a feature map of size 32x512.The dimension of this 32x512 feature map that is passed as input to the LSTM, represents 512 features per time step, where each time step represents the position for the characters that may possibly be present in the word to be recognized. Each of these timesteps contain 512 relevant features extracted by the CNN layers. There are 32 timesteps because we set the maximum length of the character sequence that can be recognized, to 32. We found that for values greater than 32 the system performed worse and the loss values were considerably higher.

The 2D LSTM which was built using 256 hidden cells, processes this feature map further, by only carrying forward the relevant information. The output of the 2D LSTM is finally mapped to a matrix of dimension 32x80, where 80 is the number of possible characters that can be recognized. This is because, apart from the 79 different characters present in our training dataset, another extra character is required for CTC operations, known as the CTC blank. Therefore, this 32x80 matrix contains the probability scores with respect to the 80 different possible entries for each timestep.

This matrix is provided as input to the CTC layer, which during training is compared with the ground truth tensor to generate a CTC loss value. This CTC loss value was the error metric that was considered for training. We used an RMSProp optimizer with a decaying learning rate that was initialized to 0.01, and a batch size of 50 for the training process. During the inference phase, the text in the image is decoded by the CTC layer using a CTC beam search decoding algorithm, which is offered as a feature in the Neural Net module of Tensorflow. This algorithm was used instead of the standard greedy best path decoding algorithm, because of which we were able to improve the accuracy of the recognized words even further.

An interactive Deep Learning based Annotation system for English handwritten text 13

Figure 7 shows the summary of our recognition model architecture.

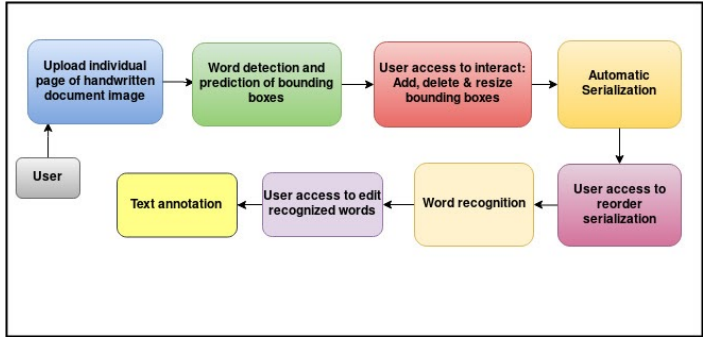

**Fig. 6** An overview of the sequence of phases in our pipeline

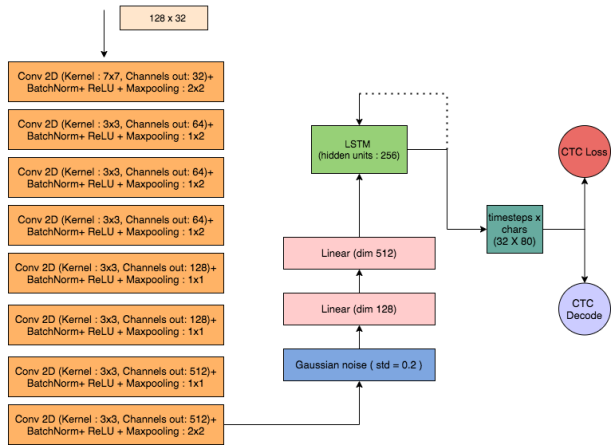

**Fig. 7** Summary of our recognition system architecture

## 5 Results and Discussions

The IAM offline dataset which was preprocessed as discussed in Section 3.2, was used to train the recognition system, with a train-valid split ratio of 95:5. Therefore, a total of 115320 words were used to train the model. In order to measure the performance of our recognition system, Character Error Rate (CER) was used as the error criterion, which is the standard among most works in this field. The CER was calculated based on the Levenshtein edit distance between the recognized word and the groundtruth word. This edit distance value between each ground truth and corresponding recognized word, was summed for all the words in the epoch, and divided by the total number of characters in all the words in the epoch, to get the final CER value. Table 1 shows the CER obtained using the two different architectures that we tried for the recognition system: one that implemented a multi-dimensional LSTM, and the other that used a single dimensional bidirectional LSTM. The significant reduction in error rate upon using a 2D LSTM justifies our choice of building a custom multidimensional LSTM instead of using a standard unidimensional, bidirectional LSTM. Table 2 further compares the results of our work with other works done in recent times that aimed to do similar tasks after being trained on the same datasets. As it can be observed, our proposed model performs better than all of them.

The decoding algorithm that we used during inference was a beam search decoding algorithm as mentioned in Section 4.4. Even though this gave much better results as compared to the greedy CTC decoder algorithm, we tried using an even better decoding algorithm known as word beam search decoding, in an attempt to further the recognition accuracy. This resulted in a small improvement in validation word accuracy. However, using a word beam search decoder limits the words recognized to those present in a dictionary/corpus that is created in the process. Thus, to make sure that our system is not limited by such constraints and performs well on words never seen before, we chose the beam search decoding algorithm over the word beam search decoding algorithm.

| Model Architecture | Character Error Rate (CER) | Epochs trained |
|---|---|---|
| CNN + Bidirectional LSTM (single dimension) + CTC | 12.36 | 110 |
| CNN + multidimensional LSTM + CTC | 9.3 | 100 |

Table 1 Comparison of the two recognition models implementing different LSTM architectures (in terms of CER)

| Work | Model | CER |
|---|---|---|
| Ingle R et al. [30] | Two Bidirectional LSTM layers | 12.8 |
| Ingle R et al. [30] | 1-D gated recurrent convolutional layers | 14.1 |
| J. Almazan et al. [31] | Kernelized Common Subspace Regression | 11.27 |
| T. Bluche [32] | Multi-layer perceptrons and Hidden Markov system | 15.6 |
| **Our work** | **CNN + multidimensional LSTM + CTC** | **9.3** |

Table 2 Comparison with other works that implemented different recognition architectures and were trained on the IAM offline dataset

To further check the robustness of our recognition system, we checked it with handwritten text that have writing styles which are harder to interpret than the text in the IAM dataset. For this we used the CVL dataset. Figure 8 shows how the system was able to pass all the phases of the pipeline successfully, and recognize the words from the CVL dataset to great accuracy, even though these words are not very legible. It is clear from our results and methodology that the proposed pipeline not only provides very impressive results for automatically annotating the data, but also ensures that the user has to put in a negligible amount of effort to ensure a near flawless annotation.

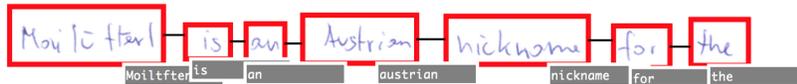

**Fig. 8** Words from the CVL dataset recognized by our recognition system.

## 6 Conclusion and Future works

After observing the burgeoning need for annotated data in the field of handwritten text recognition, we have presented a robust and innovative pipeline that carries out annotation of cursive and print handwritten English text, with a great accuracy while ensuring the least amount of human effort required. We present an annotation pipeline that uses a word detection system based on a state-of-the-art model, and combine it with a powerful custom designed recognition system. To deal with the common errors that are committed by these systems, we provide a very intuitive interactive interface which ensures the removal of a majority of the flaws with minimum effort, resulting in very fast, high quality annotated data. The potential for this pipeline is very high and has various applications. It can be used to create large amounts of custom annotated data very easily, that can be used to train systems that focus on problems such as restoring old handwritten scriptures and manuscripts, evaluating exam answer sheets, designing literary based softwares, digitizing handwritten notes and much more.

This work has potential to further be improved, by using deeper networks and larger datasets in the case of availability of powerful computational systems and hardware, which we did not have access to. Better preprocessing techniques and more powerful CTC decoding algorithms are also aspects of the work that can be focussed on for significant improvements.

Future work can focus on extending our current work to regional languages, where there is a clear lack of significant amounts of annotated data. Designing annotation pipelines for regional languages may require more sophisticated segmentation techniques and better recognition systems, and is definitely something that requires much more meticulous research. Even though there are a large number of problems that can be solved by developing systems that aim to digitize or restore documents in regional languages, one of the main limiting factors for extensive research in this field is the dearth of high quality annotated data.

# References

1. Gilloux M. (1994) Hidden Markov Models in Handwriting Recognition. In: Impedovo S. (eds) Fundamentals in Handwriting Recognition. NATO ASI Series(Series F: Computer and Systems Sciences), vol 124. Springer, Berlin, Heidelberg.
2. B. Balci, D. Saadati and D. Shiferaw, "Handwritten Text Recognition Using Deep Learning," CS231n: Convolutional Neural Networks for Visual Recognition, Stanford Uni., Course Project Report, 2017.
3. Raymond Ptucha, Felipe Petroski Such, Suhas Pillai, Frank Brockler, Vatsala Singh, and Paul Hutkowski, "Intelligent character recognition using fully convolutional neural networks" Pattern Recognition, 88:604-613, 2019
4. Sukanya, Roy, et al. "A Study on Handwriting Analysis by OCR."International Journal of Scientific and Research Publications, Volume 8, Issue 1, January 2018.
5. N.S.Manikandan and K.Ganesan. "Deep Learning Based Automatic Video Annotation Tool for Self-Driving Car." (2019).
6. Kiyokawa, Takuya & TOMOCHIKA, Keita & Takamatsu, Jun & Ogasawara, Tsukasa. (2019). Fully Automated Annotation With Noise-Masked Visual Markers for Deep-Learning-Based Object Detection. 4. 1972-1977. 10.1109/LRA.2019.2899153.
7. Tang, Jie, et al. "Automatic Semantic Annotation Using Machine Learning." Machine Learning, 2012, pp. 535–578.
8. U. Bhattacharya, R. Banerjee, S. Baral, R. De and S. K. Parui, "A Semi-automatic Annotation Scheme for Bangla Online Mixed Cursive Handwriting Samples," 2012 International Conference on Frontiers in Handwriting Recognition, Bari, 2012, pp. 680-685.
9. Stork L., Weber A., van den Herik J., Plaat A., Verbeek F., Wolstencroft K. (2019) Automated Semantic Annotation of Species Names in Handwritten Texts. In: Azzopardi L., Stein B., Fuhr N., Mayr P., Hauff C., Hiemstra D. (eds) Advances in Information Retrieval. ECIR 2019. Lecture Notes in Computer Science, vol 11437. Springer, Cham.
10. H. Q. Ung, M. K. Phan, H. T. Nguyen and M. Nakagawa, "Strategy and Tools for Collecting and Annotating Handwritten Descriptive Answers for Developing Automatic and Semi-Automatic Marking - An Initial Effort to Math," 2019 International Conference on Document Analysis and Recognition Workshops (ICDARW), Sydney, Australia, 2019, pp. 13-18.
11. V. Lavrenko, T. M. Rath and R. Manmatha, "Holistic word recognition for handwritten historical documents," First International Workshop on Document Image Analysis for Libraries, 2004. Proceedings., Palo Alto, CA, USA, 2004, pp. 278-287.
12. Memon, Jamshed, et al. "Handwritten Optical Character Recognition (OCR): A Comprehensive Systematic Literature Review (SLR)." IEEE Access, vol. 8, 2020, pp. 142642–142668.
13. Namysl, Marcin, and Iuliu Konya. "Efficient, Lexicon-Free OCR Using Deep Learning." 2019 International Conference on Document Analysis and Recognition (ICDAR), 2019.
14. C. Bartz, H. Yang and C. Meinel, "STN-OCR: A single neural network for text detection and text recognition", arXiv preprint arXiv:1707.08831, 2017.
15. Recursive Recurrent Nets with Attention Modeling for OCR in the Wild - Chen-Yu Lee, Simon Osindero 2016 IEEE Conference on Computer Vision and Pattern Recognition (CVPR) - 2016
16. Hamad, Karez & Kaya, Mehmet. (2016). A Detailed Analysis of Optical Character Recognition Technology. International Journal of Applied Mathematics, Electronics and Computers. 4. 244-244. 10.18100/ijamec.270374.

17. Scheidl, Harald, "Thesis on Handwritten Text Recognition in Historical Documents". Technische Universität Wien. 2018.
18. Graves, Alex, et al. "Multi-Dimensional Recurrent Neural Networks." Lecture Notes in Computer Science Artificial Neural Networks – ICANN 2007, 2007, pp. 549–558.
19. Wang, Kai, et al. "End-to-End Scene Text Recognition." 2011 International Conference on Computer Vision, 2011.
20. He, Pan, et al. "Single Shot Text Detector with Regional Attention." 2017 IEEE International Conference on Computer Vision (ICCV), 2017.
21. Huang, Zhihu, et al. "Text extraction in natural scenes using region-based method." Journal of Digital Information Management, vol. 12, no. 4, 2014
22. Zhou, Xinyu, et al. "EAST: An Efficient and Accurate Scene Text Detector." 2017 IEEE Conference on Computer Vision and Pattern Recognition (CVPR), 2017, doi:10.1109/cvpr.2017.283.
23. ICDAR 2013: D. Karatzas, F. Shafait, S. Uchida, M. Iwamura, L. G. i Bigorda, S. R. Mestre, J. Mas, D. F. Mota, J. A. Almazan, and L. P. de las Heras. ICDAR 2013 robust reading competition. In Proc. of ICDAR, 2013.
24. ICDAR 2015 : D. Karatzas, L. Gomez-Bigorda, A. Nicolaou, S. Ghosh, A. Bagdanov, M. Iwamura, J. Matas, L. Neumann, V. R. Chandrasekhar, S. Lu, F. Shafait, S. Uchida, and E. Valveny. ICDAR 2015 competition on robust reading. In Proc. of ICDAR, 2015
25. U. Marti and H. Bunke. The IAM-database: An English Sentence Database for Off-line Handwriting Recognition. Int. Journal on Document Analysis and Recognition, Volume 5, pages 39 - 46, 2002.
26. Florian Kleber, Stefan Fiel, Markus Diem and Robert Sablatnig, CVL-Database: An Off-line Database for Writer Retrieval, Writer Identification and Word Spotting, In Proc. of the 12th Int. Conference on Document Analysis and Recognition (ICDAR) 2013, pp. 560-564, 2013. 26
27. Vinciarelli, Alessandro, and Juergen Luettin. "A New Normalization Technique for Cursive Handwritten Words." Pattern Recognition Letters, vol. 22, no. 9, 2001, pp. 1043–1050.
28. Moysset, Bastien, and Ronaldo Messina. "Are 2D-LSTM Really Dead for Offline Text Recognition?" International Journal on Document Analysis and Recognition (IJDAR), vol. 22, no. 3, 2019, pp. 193–208., doi:10.1007/s10032-019-00325-0.
29. Connectionist temporal classification - Alex Graves, Santiago Fernández, Faustino Gomez, Jürgen Schmidhuber. Proceedings of the 23rd international conference on Machine learning - ICML '06 - 2006
30. Ingle, R. & Fujii, Yasuhisa & Deselaers, Thomas & Baccash, Jonathan & Popat, Ashok. (2019). A Scalable Handwritten Text Recognition System. 17-24. 10.1109/ICDAR.2019.00013.
31. J. Almazan, A. Gordo, A. Fornes, and E. Valveny. Word spotting and recognition with embedded attributes. IEEE Transactions on Pattern Analysis & Machine Intelligence, (12):2552–2566, 2014.
32. T. Bluche, "Deep neural networks for large vocabulary handwritten text recognition," Ph.D. dissertation, Université Paris Sud Paris XI, 2015.